\documentclass[conference]{IEEEtran}
\IEEEoverridecommandlockouts
\usepackage{cite}
\usepackage{amsmath,amssymb,amsfonts}
\usepackage{algorithmic}
\usepackage{graphicx}
\usepackage{textcomp}
\usepackage{xcolor}
\usepackage{multirow}
\usepackage{caption}
\usepackage{subcaption}
\usepackage{nicematrix}

\def\BibTeX{{\rm B\kern-.05em{\sc i\kern-.025em b}\kern-.08em
    T\kern-.1667em\lower.7ex\hbox{E}\kern-.125emX}}
\begin{document}

\title{Road Rutting Detection using Deep Learning on Images\\
}

\author{
\IEEEauthorblockN{Poonam Kumari Saha}
\IEEEauthorblockA{\textit{Center for Spatial Information Science}\\
\textit{The University of Tokyo}\\
Tokyo, Japan\\
poonamkumarisaha@g.ecc.u-tokyo.ac.jp}\\
\IEEEauthorblockN{Hiroya Maeda}
\IEEEauthorblockA{\textit{Founder, President, and CEO}\\
\textit{Urban-X Technologies, Inc.}\\
Tokyo, Japan\\
hiroya\_maeda@urbanx-tech.com}
\and
\IEEEauthorblockN{Deeksha Arya}
\IEEEauthorblockA{\textit{Center for Spatial Information Science}\\
\textit{The University of Tokyo}\\
Tokyo, Japan\\
deeksha@iis.u-tokyo.ac.jp}\\
\IEEEauthorblockN{Yoshihide Sekimoto}
\IEEEauthorblockA{\textit{Center for Spatial Information Science}\\
\textit{The University of Tokyo}\\
Tokyo, Japan\\
sekimoto@csis.u-tokyo.ac.jp}
\and
\IEEEauthorblockN{Ashutosh Kumar}
\IEEEauthorblockA{\textit{Center for Spatial Information Science}\\
\textit{The University of Tokyo}\\
Tokyo, Japan\\
ashutosh@csis.u-tokyo.ac.jp}
}
\maketitle

\begin{abstract}
Road rutting is a severe road distress that can cause premature failure of road incurring early and costly maintenance costs. Research on road damage detection using image processing techniques and deep learning are being actively conducted in the past few years. However, these researches are mostly focused on detection of cracks, potholes, and their variants. Very few research has been done on the detection of road rutting. This paper proposes a novel road rutting dataset comprising of 949 images and provides both object level and pixel level annotations. Object detection models and semantic segmentation models were deployed to detect road rutting on the proposed dataset, and quantitative and qualitative analysis of model predictions were done to evaluate model performance and identify challenges faced in the detection of road rutting using the proposed method. Object detection model YOLOX-s achieves mAP@IoU=0.5 of 61.6\% and semantic segmentation model PSPNet (Resnet-50) achieves IoU of 54.69 and accuracy of 72.67, thus providing a benchmark accuracy for similar work in future. The proposed road rutting dataset and the results of our research study will help accelerate the research on detection of road rutting using deep learning.

\end{abstract}


\section{Introduction}
\label{introduction}
Road rutting is considered a severe pavement damage \cite{JICAroadpavement2020} and is crucial in the planning of road maintenance. It affects pavement structural integrity and can cause premature failure of road surfaces. Its presence increases the possibility of hydroplaning, where water or ice accumulates in the ruts leading to the loss of grip between the pavement and the vehicle tires, as well as steering problems \cite{mamlouk2018effects}. This negatively impacts the safety of the driving public by causing undesirable vehicle vibration and vehicle instability. Therefore, there is a need to monitor road conditions and repair road rutting regularly or before it becomes severe.

Traditional methods perform inspections either manually or using specially designed vehicle equipped with sensors and laser technology, etc. These may give highly accurate results, but are very expensive to implement and maintain. High cost acts as a bottleneck in effective road maintenance especially when the budget is decreasing \cite{PavementmaintenanceinJapan}. Further, failed pavements require costly maintenance and repair which in turn cause restrictions in traffic flow. Therefore, there is a need for an easy and efficient method to detect the instances of road rutting in order to assist prompt maintenance and early rehabilitation of damaged roads. 

Recent researches have shown the potential of using  image processing techniques and deep learning in the detection of road damages (\cite{bello2014image}, \cite{katageri2016automated}, \cite{maeda2018road}, \cite{jana2022transfer}). Most of the past researches (\cite{jana2022transfer}, \cite{benmhahe2021automated}, \cite{zhang2022pavement}) and Global Road Damage Detection Challenge (GRDDC) organized as a part of the 2020 IEEE International Conference on Big Data \cite{9377790} in this direction focus on the detection of cracks, potholes, and their variants. However, detection of road rutting using deep learning still remains an open research problem. To the best of our knowledge, there is no pre-existing road rutting dataset, and presently even if it exists, the instances of road rutting are grouped with other categories such as bumps, blurring, pothole, etc. Further, its instances are remarkably less than those of other damage types. This not only affects its detection accuracy \cite{cao2020survey} but also makes it unsuitable to detect it individually as a separate category.

This paper addresses these needs and makes the following contributions in this area:
\begin{enumerate}
    \item Proposes a road rutting data set containing 949 images collected from heterogeneous sources.
    \item Provides data annotation in both object level bounding box and pixel level formats.
    \item Proposes object detection as well as semantic segmentation models capable of detecting road rutting and presents a benchmark for its detection accuracy.
    \item Comprehensively analyses the challenges faced in the detection of road rutting using deep learning.
\end{enumerate}

As compared to traditional methods, the proposed method requires less energy, manpower, and financial resources. It is easy to standardize and reproduce. Addition of more training data and improvement in deep learning techniques in future will only make this method better and stronger.

The rest of the paper is organized as follows. In section~\ref{sec:related_works}, related works are discussed. Details of the proposed dataset and methodology are explained in section~\ref{sec:methodology}. Results are provided in section~\ref{sec:results}. Discussion on the results is provided in section~\ref{sec:discussions}. Finally, section~\ref{sec:conclusions} concludes the paper.

\section{Related Works}
\label{sec:related_works}
This section describes existing methods and research works related to the detection of road rutting, ranging from traditional methods to modern machine learning based methods. Further, image processing and deep learning based methods are focused and thereafter, availability of road rutting dataset, object detection models, and image semantic segmentation models are discussed.

\subsection{Traditional Methods of Road Rutting Detection}
Many traditional methods are based on the measurement of road rut depth. This is done either manually or using equipment such as laser scanning technology \cite{PavementmaintenanceinJapan}, profilometers equipped with infrared sensors, rut bar collection system, ultrasonic technology \cite{rajab2008application}, etc. Some studies also deploy a photographic data collection system using a camera and a strobe and use Pavement Distress Analysis System (PADIAS) \cite{simpson2001measurement} for doing measurements. An accurate and cost-effective autonomous pavement rutting measurement by fusing a high speed-shot camera and a linear laser has been proposed in \cite{arezoumand2021automatic}.

Although these quantitative inspections are highly accurate, it is considerably expensive to conduct such comprehensive inspections especially for small municipalities that lack the required financial resources. Further, these methods are very specific to the detection of road rutting and cannot be generalized to detect other road damages. Furthermore, the use of specialized vehicles with high-precision sensors to collect road condition data requires dedicated infrastructure and manpower which makes it expensive \cite{du2021application}. High cost can discourage municipal road maintenance bodies to perform frequent monitoring trips given the limited availability of allocated funds and manpower.

\subsection{Predictive Models using Machine Learning Techniques}
Recently proposed methods for road rutting detection involve collection of road rutting data such as depth, width, radius of curvature of ruts; temperature, pavement age, and other factors related to the permanent deformation of the section, etc. The data so obtained is then used to compute some of the pavement performance indices or establish relationship amongst different factors using machine learning to develop predictive models. Authors in \cite{taddesse2013intelligent} have developed an intelligent pavement rutting prediction model by applying multiple linear regression (MLR) and artificial neural network (ANN) techniques on Norwegian national road databank (NVDB). A generic pavement rutting model has been developed in \cite{haddad2021use} by deploying deep neural network techniques (DNN) on data extracted from the Long-Term Pavement Performance (LTPP) database.

Since different methods and different combinations of factors are considered at different places and these analyses being done are very specific to the location, it is challenging to generalize these practices across the country or globe.

The aforesaid limitations reiterate the need for an automated damage detection technology that requires less resources, and is easy to implement, standardize, and reproduce.

\subsection{Road Damage Detection using Image Processing and Deep Learning}
Recent researches have shown impressive results on road damage detection using image processing and deep learning. Many researches have been done to detect road cracks and potholes such as in \cite{zhang2022pavement}, \cite{subirats2006automation}, \cite{mandal2015automated}, \cite{vigneshwar2016detection}, \cite{bhatia2019convolutional}, etc. Authors in \cite{ciaparrone2018deep} have used fully convolutional network (FCN) and faster region-based convolutional neural networks (Faster R-CNN) for road damage detection in Naples (Italy), but their dataset contains only one instance of road rutting out of a total of 8736 road damage instances. Authors in \cite{maeda2018road} have used single shot multi-box detector (SSD) with Inception V2 and MobileNet for road damage detection in their proposed dataset, RDD2018, for Japan, but their dataset groups rutting with other road damages such as blur, separation, and pothole. Their model resulted in low recall of this category which was attributed to less number of training data. The dataset RDD2018 was extended to include images from multiple countries in RDD2020 \cite{arya2021deep}. However, the extended dataset does not include road rutting. Further, recently released multi-country road damage dataset RDD2022 \cite{arya2022rdd2022} deals with only cracks and potholes.

\subsection{Available Road Rutting Dataset}
The road damage datasets made available by the works of \cite{maeda2018road}, \cite{subirats2006automation}, \cite{arya2021deep}, \cite{vickers2017animal}, \cite{tree2012automatic}, etc. mainly contains instances of cracks, potholes, and their variants. In contrast, the road damage dataset proposed in \cite{hadjidemetriou2019vision} contains 263 images of road rutting. However, these images are captured perpendicularly above road surfaces and are not wide-view images obtained from vehicle mounted cameras or smartphones.

\subsection{Object Detection Models}
An object detection model finds the number of objects in an image and estimates bounding box coordinates for each object along with its category. In other words, it outputs the class probabilities and bounding boxes around the objects on the input image fed to the network.

Object detection methods are broadly classified into two categories: two-stage detectors and one-stage detectors. A two-stage detector detects objects by first making a region proposal of possible locations of objects and then classifies the objects based on features extracted from the proposed region and the regression of rectangular bounding box coordinates. Faster R-CNN \cite{ren2015faster} and Mask R-CNN \cite{he2017mask} are two-stage detectors. A one-stage detector considers object detection as a single-stage or fully regression problem without any region proposals. SSD \cite{liu2016ssd}, RetinaNet \cite{lin2017feature}, and You Look Only Once (YOLO) \cite{redmon2016you} and its variants are one-stage detectors.

Since YOLO achieves comparable mean average precision (mAP) on the benchmark Micorsoft Common Objects in Context (COCO) \cite{lin2014microsoft} dataset while running faster than other object detection models, we primarily focus on four recent versions of YOLO for object detection in our study: YOLOv4 \cite{bochkovskiy2020yolov4}, YOLOv5 \cite{glennjocher20226222936}, YOLOv6 \cite{yolov6}, and YOLOX \cite{https://doi.org/10.48550/arxiv.2107.08430}. 

\subsection{Image Semantic Segmentation Models}
It is a form of pixel level prediction as it assigns a class label to every pixel in an image. It classifies a certain class of image and separates it from the rest of the image classes by overlaying it with a segmentation mask. It outputs the bounding box around and segmentation mask on the object when an input image is fed to the network. These models have been used in our study to analyze if it affects  the detection accuracy of road rutting by using only segmentation mask in order to avoid the additional information that gets enclosed while using bounding box in object detection model. The image semantic segmentation models used in the current study are PSPNet (Resnet-50) \cite{zhao2017pspnet} and DeepLabv3+ (Resnet-101) \cite{deeplabv3plus2018}.

\section{Methodology}
\label{sec:methodology}

The methodology consists of four steps: data collection and dataset preparation, annotation of data, performing object detection and image semantic segmentation to determine the feasibility of detection of road rutting, and quantitative and qualitative evaluation of results and analysis of possible challenges faced in detection.

\subsection{Data Collection}
In our research, a novel road rutting dataset has been developed from heterogeneous sources. Following three sources are considered:

\begin{itemize}
    \item Images extracted from the videos captured by on-board vehicle smartphone or cameras around Susono \cite{kumar2021citywide}, Hida, and Maebashi city of Japan.
    \item Images obtained from Mapillary Street-level Sequences Dataset of Tokyo, Japan \cite{warburg2020mapillary}.
    \item Images obtained from web scraping and captured by smartphone in Meguro ward of Japan.
\end{itemize}

Images from Susono city are in PNG format with a resolution of $1280 \times 720$, those from Hida and Maebashi city  are in PNG format with a resolution of $1920 \times 1080$, those from Mapillary Street-level Sequences Dataset of Tokyo are in JPG format with a resolution of $640 \times 360$, and those obtained from web scraping and captured by smartphone are in JPG format with different resolutions. Details of the dataset are shown in Table \ref{road-rutting-dataset}.

\begin{table*}[htbp]
\caption{Details of Road Rutting Dataset}
\begin{center}
\begin{tabular}{|c|c|c|c|c|}
\hline
\textbf{Sources} & \textbf{Locations} & \textbf{Image Size} & \textbf{Image Type} &  {\textbf{No. of Images}} \\
\hline
\multirow{2}{5cm}{Images extracted from videos captured by on-board vehicle cameras or smartphones} & Susono City & $1280 \times 720$ & PNG & $34$\\
\cline{2-5}
& Maebashi and Hida City & $1920 \times 1080$ & PNG & $240$\\
\hline
Mapillary Street-level Sequences Dataset & Tokyo & $640 \times 360$ & JPG & $575$\\
\hline
Web scraping and captured by smartphone & Meguro and others & variable & JPG & $100$\\
\hline
\end{tabular}
\label{road-rutting-dataset}
\end{center}
\end{table*}

\begin{figure}
     \centering
     \begin{subfigure}[t]{0.24\textwidth}
         \centering
         \includegraphics[width=\linewidth]{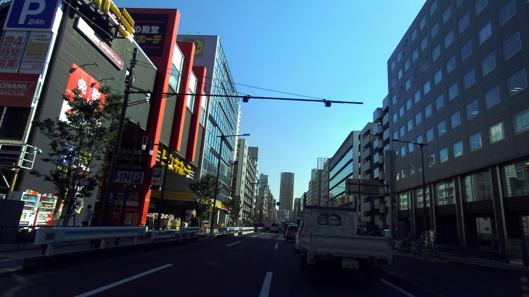}
         \subcaption{}
     \end{subfigure}
     \begin{subfigure}[t]{0.24\textwidth}
         \centering
         \includegraphics[width=\linewidth]{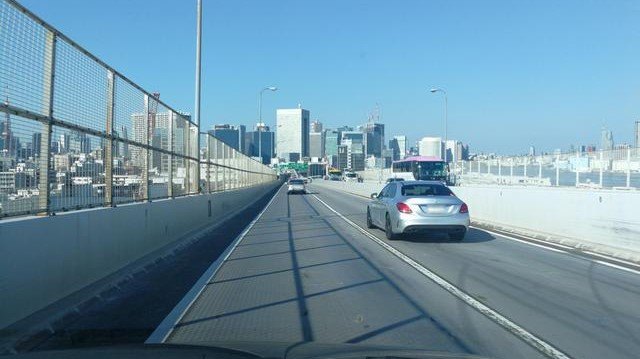}
         \subcaption{}
     \end{subfigure}
     \begin{subfigure}[t]{0.24\textwidth}
         \centering
         \includegraphics[width=\linewidth]{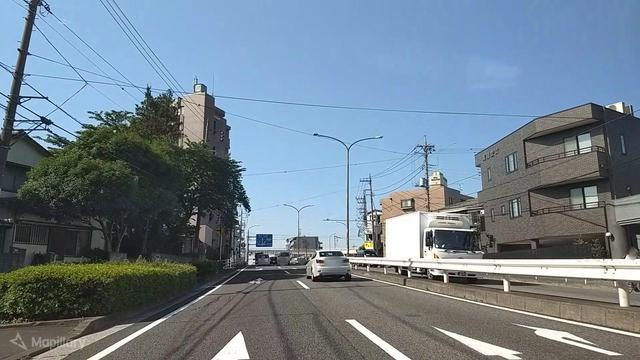}
         \subcaption{}
     \end{subfigure}
     \begin{subfigure}[t]{0.24\textwidth}
         \centering
         \includegraphics[width=\linewidth]{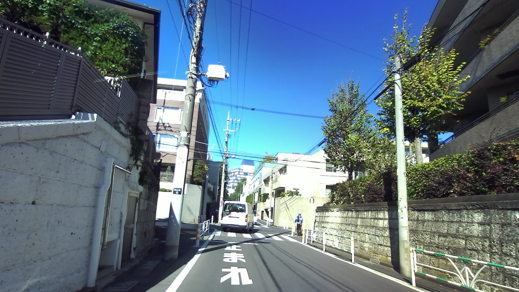}
         \subcaption{}
     \end{subfigure}
     
        \caption{Samples of discarded images: (a) and (b) contain dark and undesirable shadows, while (c) and (d) contain shadows of overhead electric wires}
        \label{fig:discarded-images}
\end{figure}

The videos have been collected in March 2021 and between March and April 2022 under varying weather and lightning conditions. A dataset of 949 images was finally created after careful selection from over 100,000 images. The selection criteria involves discarding of following types of images:

\begin{itemize}
    \item Blurred images
    \item Images with dark shadows
    \item Images that don't contain significant portions of the road
    \item Images that don't contain road rutting
    \item Images containing shadows of overhead electric wires (The reason for this is explained in section~\ref{sec:discussions}).
\end{itemize}
Samples of discarded images are illustrated in Fig. \ref{fig:discarded-images}.

\subsection{Data Annotation}
Each image has been manually annotated in a semi-automatic way to provide annotations in both object level and pixel level formats.

In the object level annotation, bounding boxes have been made using an annotation tool labelImg \cite{tzutalin2015labelimg}. The annotations are saved in YOLO labelling format which is a .txt file. This format contains information about the class and about each bounding box which is represented by four values [x\_center, y\_center, width, height] in the normalized form. At first, 308 images were annotated manually which were used to train a YOLOv4 model. This model was then used incrementally to annotate images from each subsequent sources. The annotations were checked manually using the labelImg tool, and this process was repeated.

In the pixel level annotation, semantic segmentation masks have been created using data labeling tool called Label Studio \cite{LabelStudio}. ``Semantic Segmentation with Masks" was used. This uses a brush to draw region on the image. The masks were exported in PNG format. This was used in training the semantic segmentation models in our study. In addition to masks using brush, masks using polygons have also been considered which saves the annotations in COCO format which is a .json file.

\begin{figure*}
     \centering
     \begin{tabular}[width=\textwidth]{c|c|c}
        \includegraphics[width=.31\linewidth]{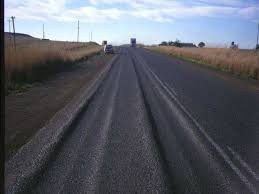}& \includegraphics[width=.31\linewidth]{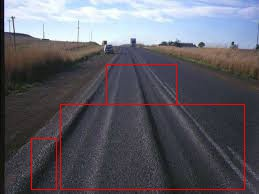}& \includegraphics[width=.31\linewidth]{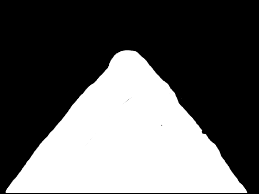}\\
        \hline
        \includegraphics[width=.31\linewidth]{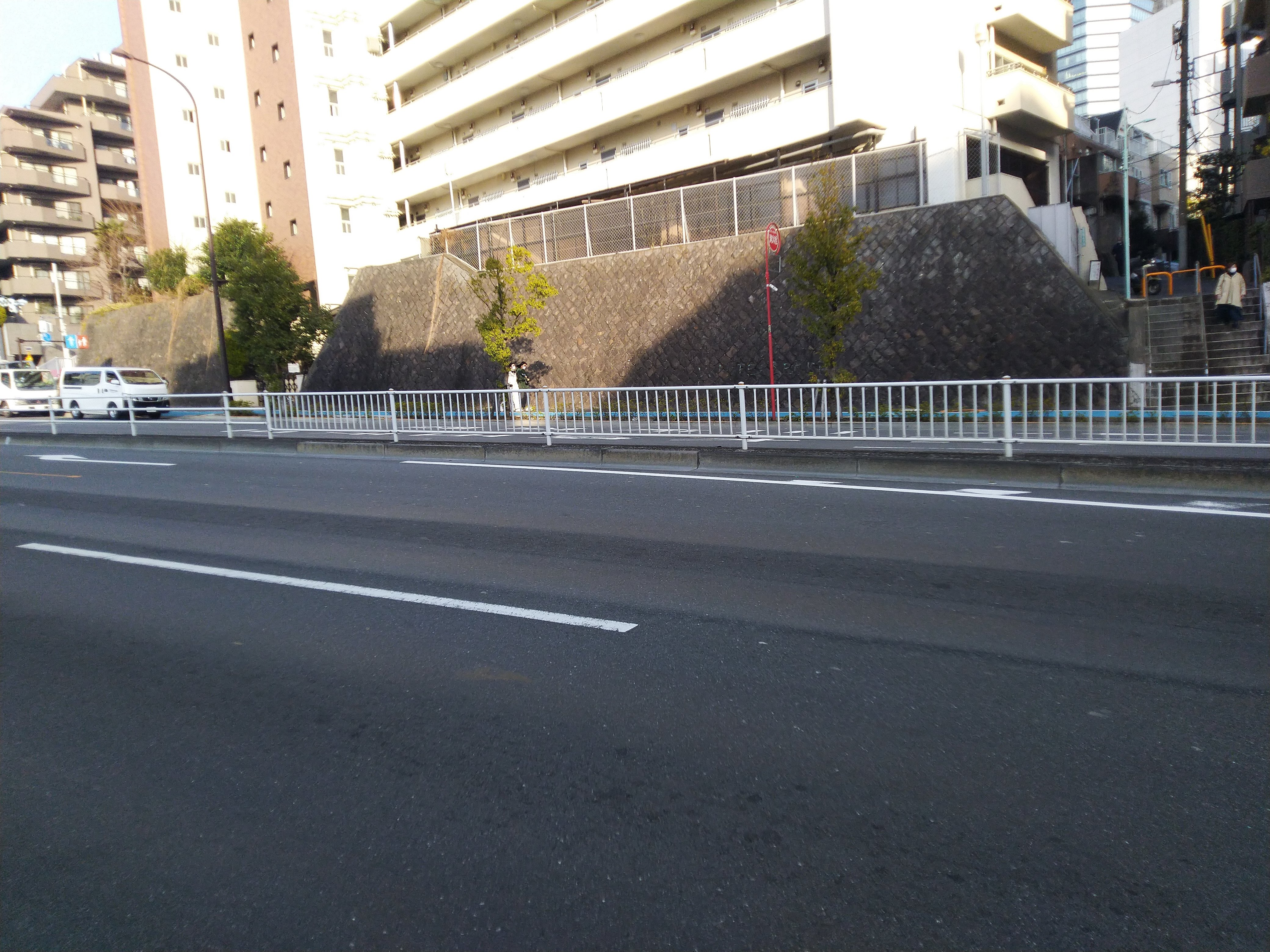}& \includegraphics[width=.31\linewidth]{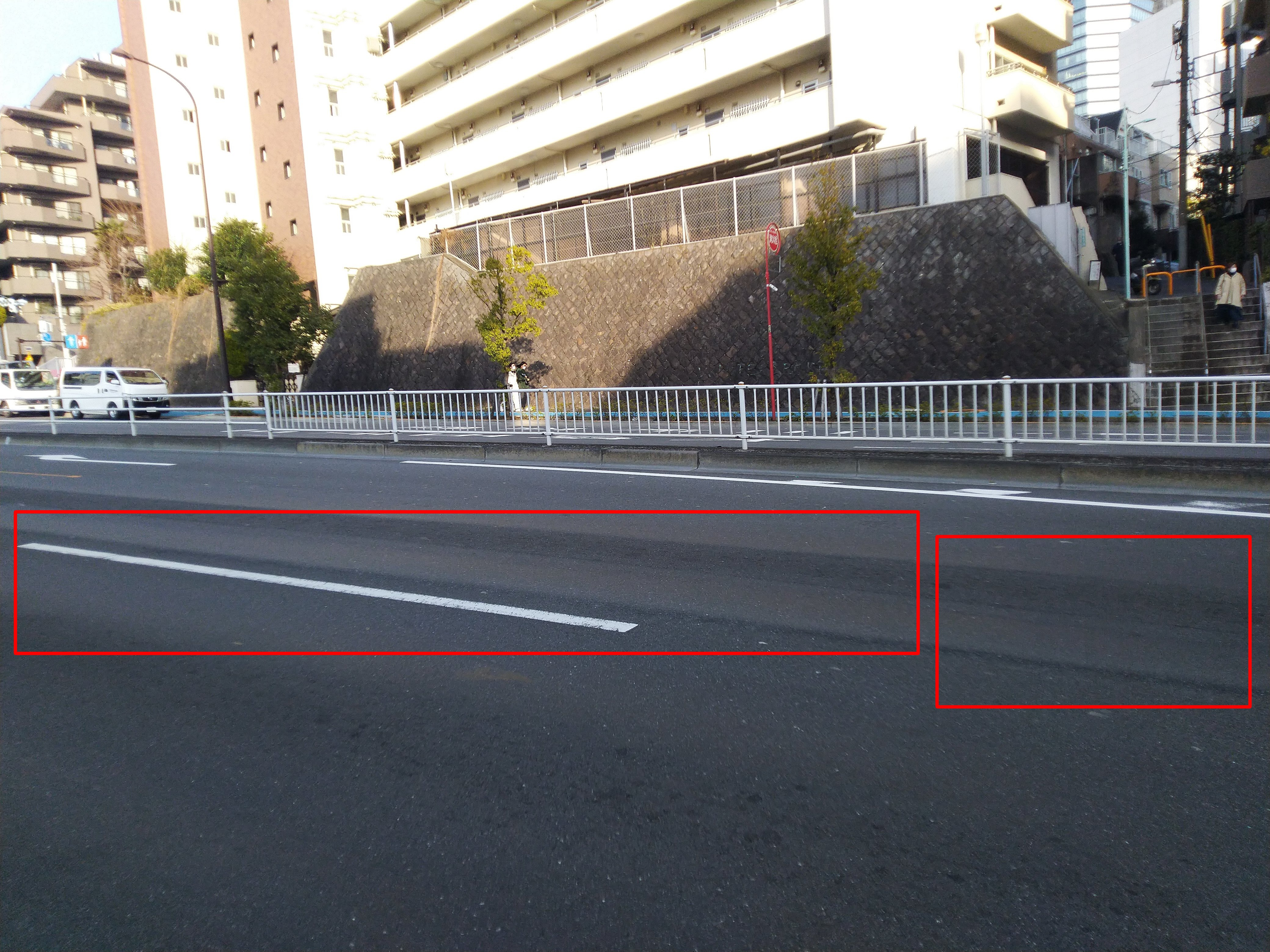}& \includegraphics[width=.31\linewidth]{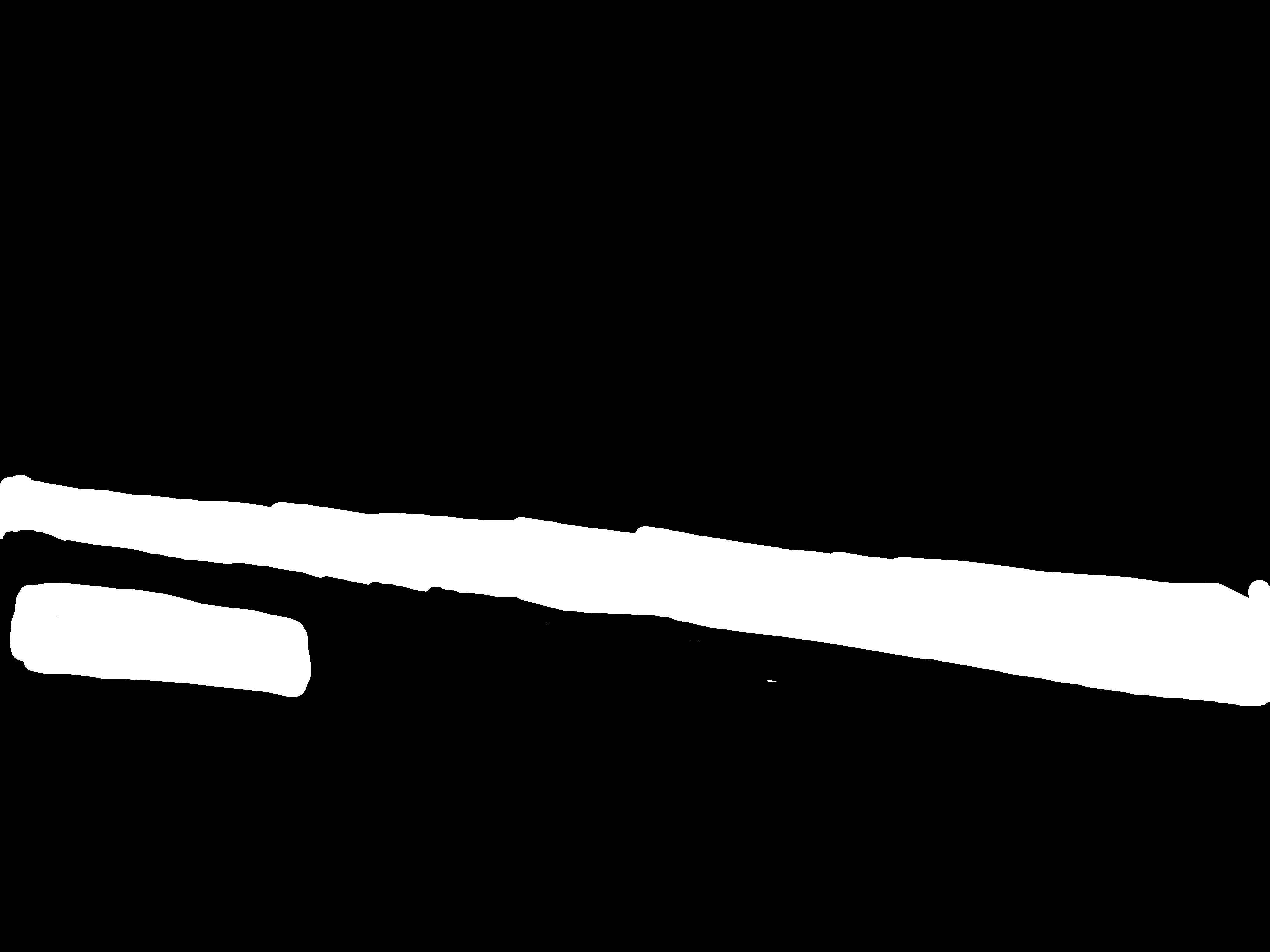}\\
        (a) Original images & (b) Images with object level annotations & (c) Images with pixel level annotations
     \end{tabular}
        \caption{Data Annotation}
        \label{fig:data-annotation}
\end{figure*}

Samples of some images along with object level bounding box and pixel level mask annotation are illustrated in Fig. \ref{fig:data-annotation}.

\subsection{Data Statistics}
The proposed road rutting dataset has 949 images containing 904 instances of rutting. The entire dataset is randomly split in the ratio of 85:15. A set of 806 images was used for training and that of 143 images was used for testing.

\begin{figure}
     \centering
     \begin{subfigure}[t]{0.24\textwidth}
         \centering
         \includegraphics[width=\linewidth]{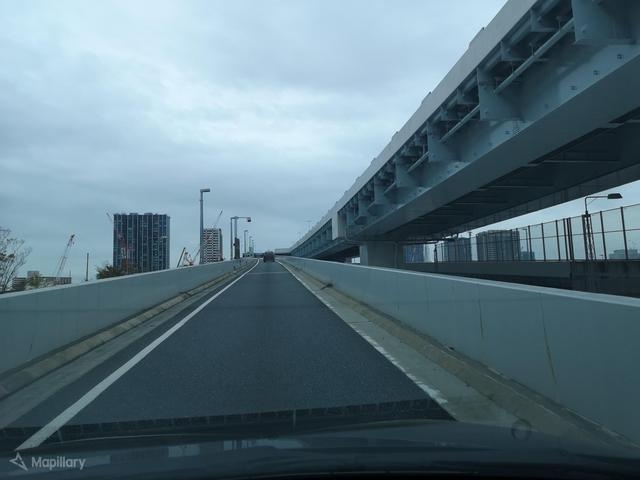}
         \subcaption{}
     \end{subfigure}
     \begin{subfigure}[t]{0.24\textwidth}
         \centering
         \includegraphics[width=\linewidth]{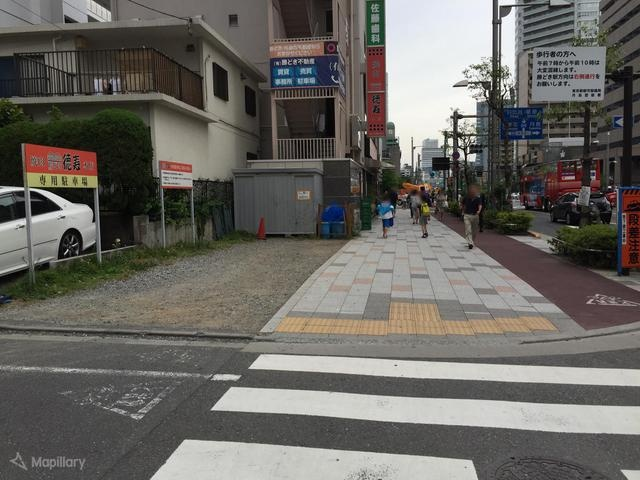}
         \subcaption{}
     \end{subfigure}
     \begin{subfigure}[t]{0.24\textwidth}
         \centering
         \includegraphics[width=\linewidth]{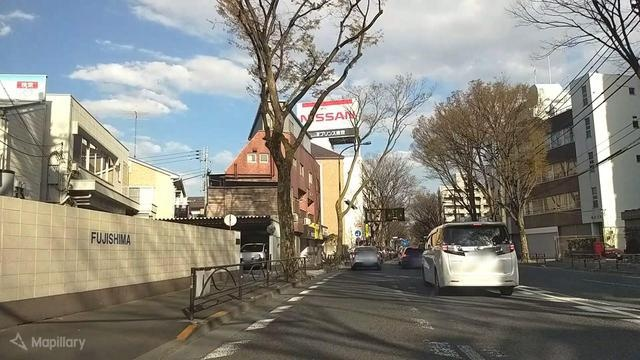}
         \subcaption{}
     \end{subfigure}
     \begin{subfigure}[t]{0.24\textwidth}
         \centering
         \includegraphics[width=\linewidth]{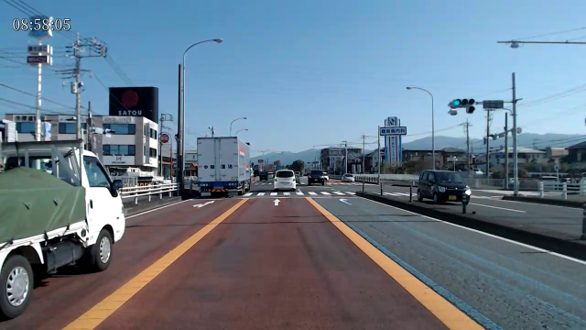}
         \subcaption{}
     \end{subfigure}
     
        \caption{Images of negative samples: (a) contains corrugated surfaces in walls nearby road surface, (b) contains zebra crossing and sidewalks, (c) contains shadows of tree, and (d) contain colorful patches on road surfaces}
        \label{fig:negative_samples}
\end{figure}

During the initial training, it was observed that model made false predictions in images containing zebra crossing, sidewalks, having corrugated surfaces near the road, and having shadows of trees or colorful patch work on road surfaces. Therefore, negative samples without bounded box were included in the dataset to help the model learn better. This is in accordance with instructions on improving detection accuracy as per \cite{bochkovskiy2020yolov4}. Consequently, images which don't contain rutting have been included in the dataset. Some images of negative samples are shown in Fig. \ref{fig:negative_samples}.

\subsection{Road Rutting Detection using Object Detection Model}
The study involves training and evaluation of object detection models using YOLOv4, YOLOv5, YOLOv6, and YOLOX. Their pre-trained weights trained on the COCO dataset were utilized to train the requisite road rutting detection models using transfer learning. The models were fine-tuned by exploring several hyperparameters such as image size, batch size, number of epochs.  The models have been trained on image size of $416 \times 416$ and evaluation has been done on confidence threshold of 0.25 and IoU threshold of 0.5 as suggested by Pascal VOC Challenge \cite{hoiem2009pascal} on the test dataset. The model weights with best mAP on the test dataset were chosen for all future evaluation purposes. 

Further, the technique of image augmentation using albumentations \cite{info11020125} was explored with the objective to improve the detection accuracy of YOLOv4 model by increasing the size of the training dataset. Horizontal flip, random brightness contrast, Gaussian noise, RGB shift, sharpen, etc. were used in accordance with techniques mentioned in \cite{arya2021deep}. However, vertical flip was not used as images with vertically flipped road will not appear in the original dataset while taking videos or images using dashboard cameras on vehicles. For the augmented dataset, mean Average Precision at Intersection over Union 0.5 (mAP@IOU=0.5) for YOLOv4 was found to be 32.2\% which is slightly lesser than the mAP@IOU=0.5 of 33.7\% obtained by YOLOv4 on dataset without image augmentation. For YOLOv5, Test Time Augmentation (TTA) was performed. The mAP@IoU=0.5 improved from 34.4\% for YOLOv5 model without TTA to only 34.7\% for YOLOv5 model with TTA. Since, results using augmentation techniques didn't provide significant improvements, it was not explored further.

\subsection{Road Rutting Detection using Image Semantic Segmentation Model}
Another aspect that arise is if the detection accuracy of road rutting can be improved by using segmentation masks instead of bounding boxes. The reason for considering this aspect stems from our preliminary visual analysis of the results of object detection based models. The use of bounding box while annotating road rutting in an image can include additional information of surrounding along with the rutting instance. The use of multiple smaller bounding boxes instead of a single larger bounding box can help address this challenge to some extent, but in doing so, some smaller portion of the rutting instance can itself get excluded. However, the use of segmentation masks can ensure that no additional information of surrounding is included in the annotation.

This study involves training and evaluation of image semantic segmentation models using PSPNet (Resnet-50) and DeepLabV3+ (Resnet-101). For details, readers may refer to OpenMMLab’s MMSegmentation, which is an open source semantic segmentation toolbox \cite{mmseg2020} based on PyTorch.

\section{Results}
\label{sec:results}
\begin{figure*}
     \centering
     \begin{tabular}[width=\textwidth]{c}
        \includegraphics[width=\linewidth]{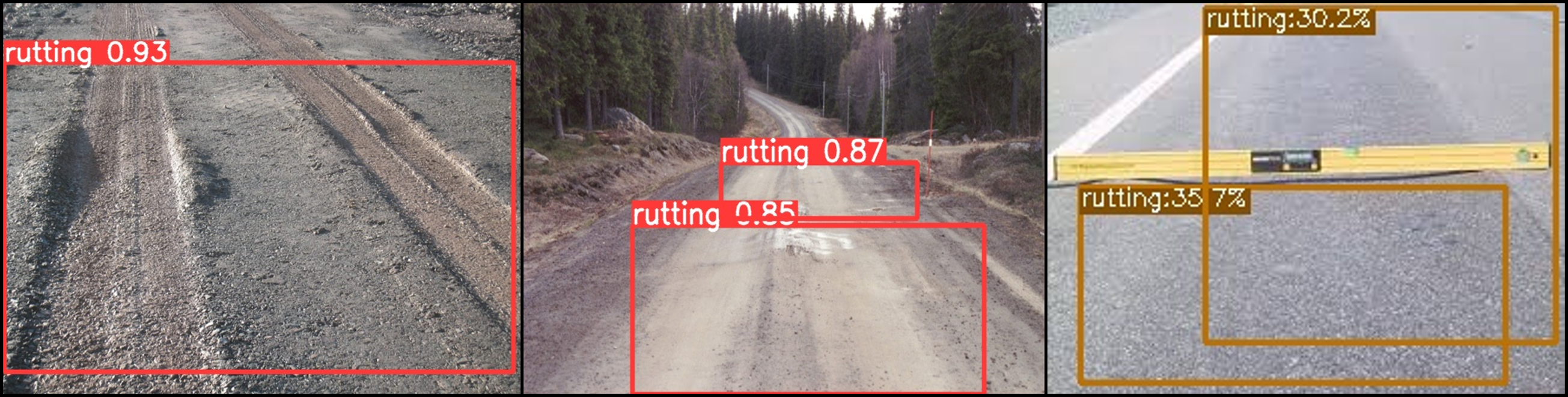}  \\
        \includegraphics[width=\linewidth]{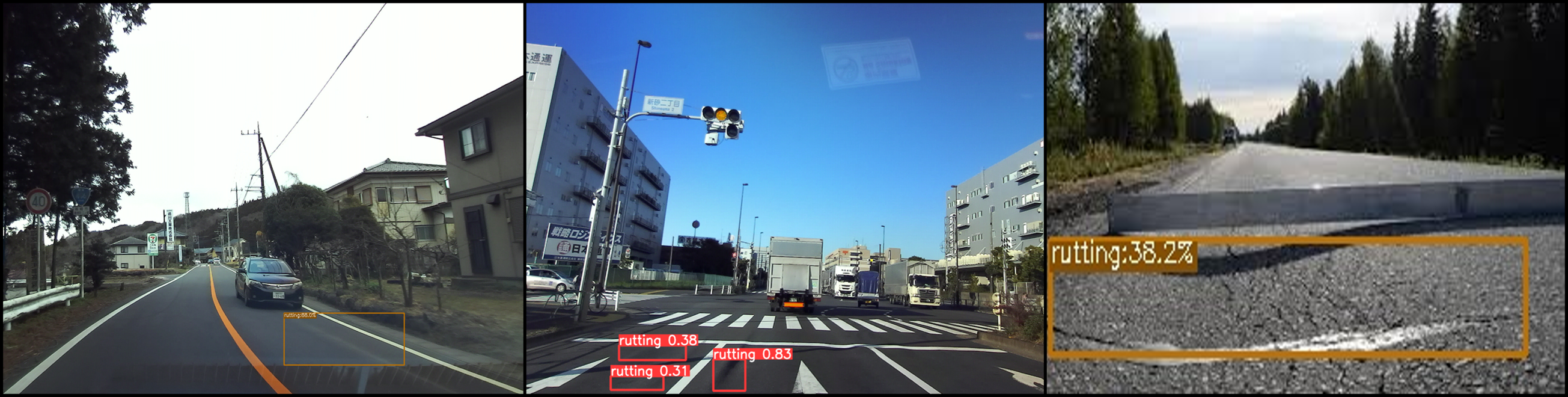} \\
        \includegraphics[width=\linewidth]{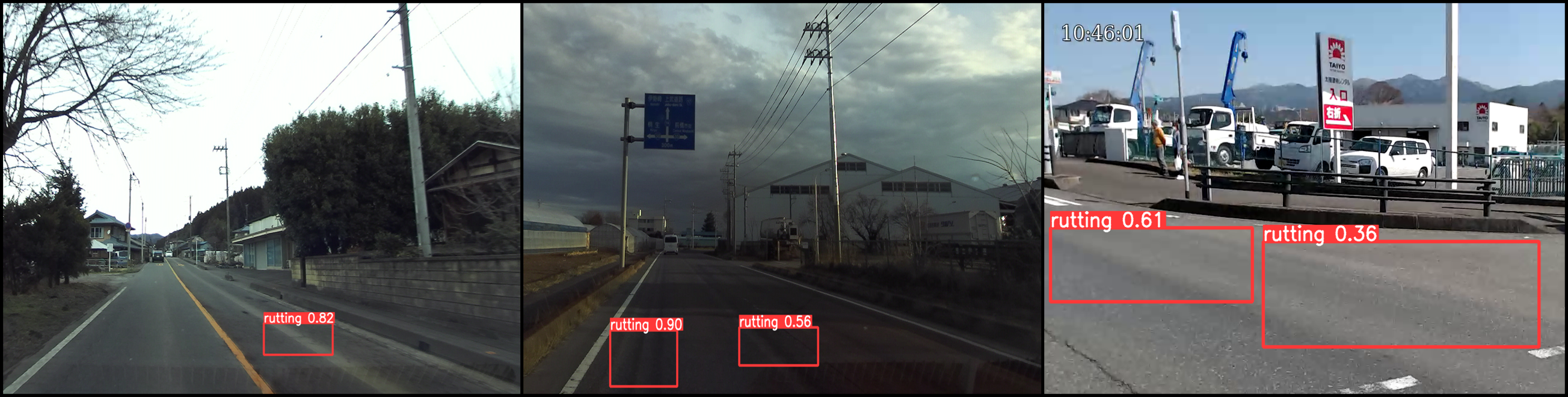}\\
     \end{tabular}
        \caption{Detected samples by object detection models}
        \label{fig:predicted-images-od}
\end{figure*}

\begin{figure*}
     \centering
     \begin{tabular}[width=\textwidth]{c}
        \includegraphics[width=\linewidth]{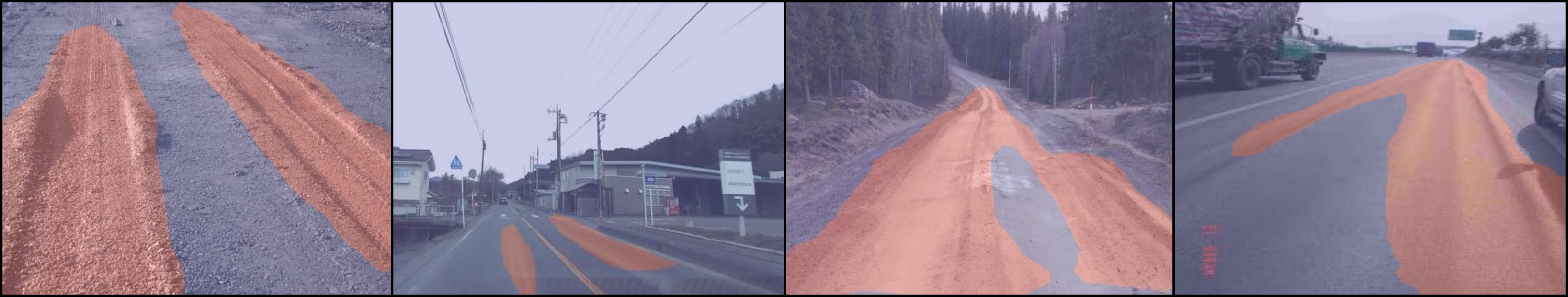} 
     \end{tabular}
        \caption{Detected samples by semantic segmentation models}
        \label{fig:predicted-images-seg}
\end{figure*}

In our experiment, training of the deep learning models was performed on AWS instance type named g4dn.xlarge running on the Ubuntu 18.04 operating system. It has one NVIDIA-T4 GPU with 16 GiB GPU memory. The detected samples using object detection models are illustrated in Fig. \ref{fig:predicted-images-od} and those using semantic segmentation models are shown in Fig. \ref{fig:predicted-images-seg}.

\subsection{Performance of Different Object Detection Models}
This study considers mAP@IoU=0.5 as the quantitative measure to compare the performance of different object detection models. This measure has the advantage that it provides a single value for each object detection model to compare its performance with other object detection models. Higher the value of mAP, better is the performance of the model. Table \ref{objectDetection-result} presents mAP obtained by different object detection models used in the study.

\begin{table}[htbp]
\caption{Performance of Object Detection Models used in the study}
\begin{center}
\begin{tabular}{|c|c|}
\hline
\textbf{Model name} & {\textbf{mAP@IoU=0.5 (\%)}}\\
\hline
YOLOv4  &  33.7 \\
\hline
YOLOv5-l & 34.4 \\
\hline
YOLOv6-s (finetune) & 26.9 \\
\hline
YOLOv6-s & 20.3 \\
\hline
YOLOX-s & {\textbf{61.6}} \\
\hline
YOLOX-m & 59.7 \\
\hline
YOLOX-l & 55.9 \\
\hline
\end{tabular}
\label{objectDetection-result}
\end{center}
\end{table}

YOLOX achieved the maximum mAP value amongst the trained models. And amongst the variants of YOLOX, YOLOX-s achieved the highest mAP value.

\subsection{Performance of Different Image Semantic Segmentation Models}
This study considers Intersection over Union (IoU) and Pixel Accuracy (Acc) as the quantitative measures to compare the performance of one semantic segmentation model with that of other. Both are the commonly used evaluation metrics for comparison purpose. IoU, also referred as Jaccard Index, is essentially a method to quantify the percent overlap between the target mask and prediction mask. The value of IoU is from 0 and 1, and higher IoU means higher overlapping between predicted boxes and ground truth boxes and therefore, represents higher localization accuracy. On the other hand, pixel accuracy reports the percent of pixels in the image which are correctly classified. IoU is considered to be better than pixel accuracy as the latter is not suitable when one class overpowers the other. In the present case, background and rutting become two classes and in most images, background accounts for larger area than rutting does. Therefore, the semantic segmentation model with greater IoU will be considered better.

Table \ref{semantic-result} lists the result of semantic segmentation models used in this study. Both the semantic segmentation models used in the study achieve IoU above 52 and accuracy above 72.

\begin{table}[htbp]
\caption{Performance of Semantic Segmentation Models used in the study}
\begin{center}
\begin{tabular}{|c|c|c|}
\hline
\textbf{Model}&\multicolumn{2}{|c|}{\textbf{Rutting}} \\
\cline{2-3} 
\textbf{name} & \textbf{\textit{IoU}}& \textbf{\textit{Acc}} \\
\hline
PSPNet (Resnet-50)  &  54.69    &   72.67 \\
\hline
DeepLabV3+ (Resnet-101) & 52.97 &   73.95 \\
\hline
\end{tabular}
\label{semantic-result}
\end{center}
\end{table}

\subsection{Visual Analysis of Predicted labels and Masks}
Visual or qualitative analysis of the ground truth and predicted labels and masks was performed with the following two objectives:
\begin{enumerate}
    \item To evaluate and compare the performance of the object detection model and semantic segmentation model as these are two different approaches of deep learning to identify and localize the object in the image, and these don’t have a common evaluation metrics to compare their performances.
    \item To comprehensively analyze the predictions and identify the challenges in performing road rutting detection.
\end{enumerate}

\begin{figure}
     \centering
     \begin{subfigure}[t]{0.24\textwidth}
         \centering
         \includegraphics[width=\linewidth]{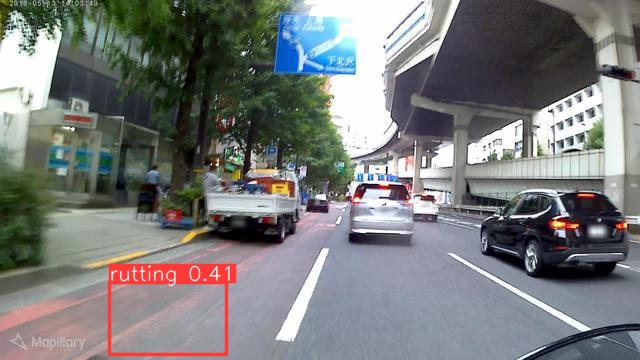}
         \subcaption{colorful patch}
     \end{subfigure}
     \begin{subfigure}[t]{0.24\textwidth}
         \centering
         \includegraphics[width=\linewidth]{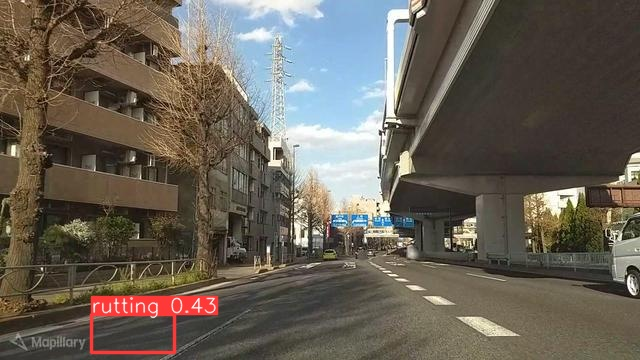}
         \subcaption{tree shadows}
     \end{subfigure}
     \begin{subfigure}[t]{0.24\textwidth}
         \centering
         \includegraphics[width=\linewidth]{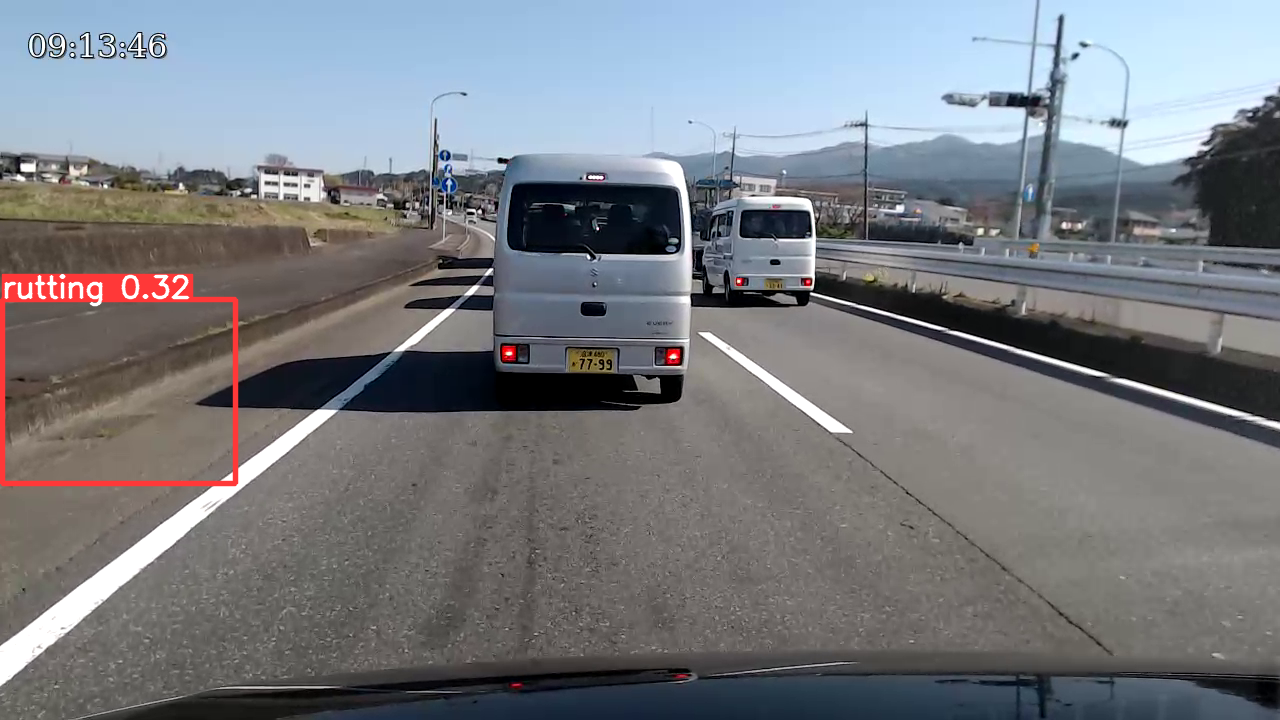}
         \subcaption{nearby road surfaces}
     \end{subfigure}
     \begin{subfigure}[t]{0.24\textwidth}
         \centering
         \includegraphics[width=\linewidth]{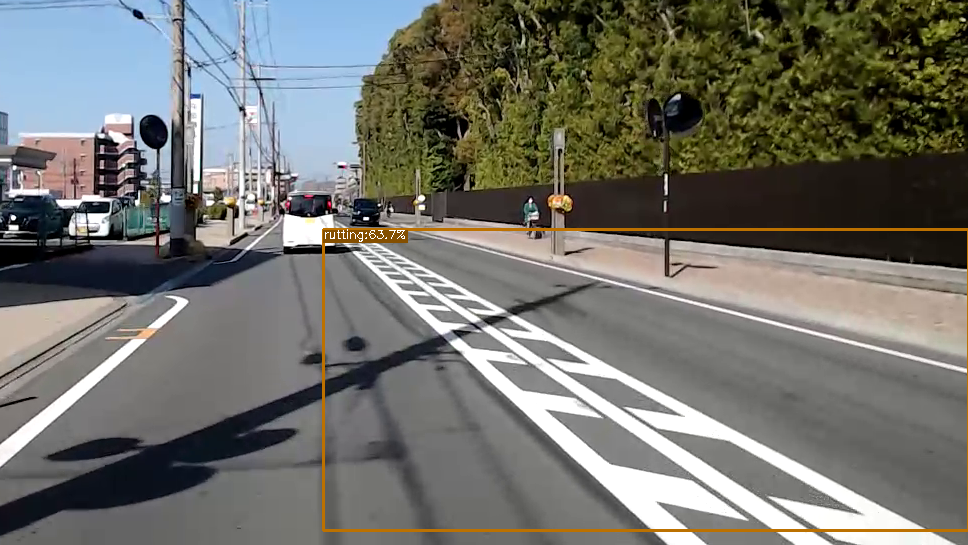}
         \subcaption{shadows of overhead electric wires}
     \end{subfigure}
  
        \caption{Samples of false prediction by object detection models}
        \label{fig:wrong-predictions-od}
\end{figure}

\begin{figure}
     \centering
     \begin{subfigure}[t]{0.24\textwidth}
         \centering
         \includegraphics[width=\linewidth]{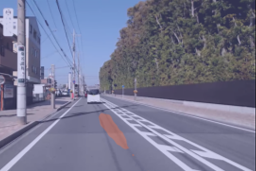}
         \subcaption{}
     \end{subfigure}
     \begin{subfigure}[t]{0.24\textwidth}
         \centering
         \includegraphics[width=\linewidth]{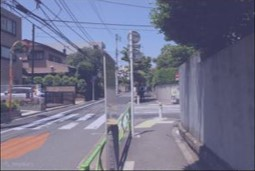}
         \subcaption{}
     \end{subfigure}
        \caption{Images of false prediction by semantic segmentation models on shadows on overhead electric wires are shown on (a) and (b)}
        \label{fig:wrong-predictions-seg}
\end{figure}

Following observations were made during the visual or qualitative analysis:

\begin{itemize}
    \item It was found on visual checking of the predicted labels or masks that all the studied models were able to correctly predict the road rutting instances on most images.
    \item The inclusion of negative samples improved the model performance. All the models correctly identified images containing zebra-crossing and side walks as non-rutting.
    \item All the object detection models except YOLOv5 incorrectly predicted rutting on some of the images with red colorful patches on road surfaces. YOLOv6 also made false predictions on some images containing shadows of trees on road and nearby road surfaces having difference in heights. In most cases, all the models incorrectly predicted dark shadows of overhead electric wires which are longitudinal to the road as rutting. The samples of wrong predictions by object detection models are shown in Fig. \ref{fig:wrong-predictions-od} and those by semantic segmentation models are shown in Fig. \ref{fig:wrong-predictions-seg}.
    \item Unlike object detection models, semantic segmentation models made correct predictions on above cases except for shadows of overhead electric wires.
    \item It was noted that the bounding boxes predicted by YOLOX were larger in size and had higher confidence score as compared to those of other object detection models used in the study.
    
\end{itemize}

     

\section{discussion}
\label{sec:discussions}
The initial success achieved by these training models in the detection of road rutting using image dataset strongly suggests that the deep learning based detection models can be used in the detection of road rutting, and it will work even more efficiently when trained on a significantly larger training dataset. More training data will improve the accuracy of prediction \cite{du2021pavement}.

Further, there cannot be a single correct bounding box or image mask for road rutting in the image. The labels or masks predicted by both object detection and semantic segmentation models are acceptable when a comprehensive survey for monitoring road conditions is required. However, since these models compares the predicted labels or image masks with the information in the ground truth for evaluation for model performance, some of the predicted labels or image masks, although acceptable, get marked as false examples. Presence of many such images in the dataset can result in the low performance of the models. This was also pointed out in \cite{arya2021deep}. 

Visual or qualitative analysis of prediction of both type of deep learning models shows that there are high chances of false detection of the dark shadows of overhead electric wires on road surfaces which are particularly longitudinal to the road as rutting. One possible reason can be that these shadows create an illusion of road rutting when although road rutting is not present on the road. Similarly, alternate worn out surfaces due to wheel path movement and fresh surfaces on the road can also pose a difficult challenge in road rutting detection. Again, if there are nearby surfaces around road which contains corrugated surfaces, grooves or some kind of depressions, then these may also get detected as road rutting. There are chances for shadows of trees on road surfaces to get falsely predicted as road rutting. These problems can be addressed by training models by including more such images and annotating it as not rutting. The use of road segmentation can also improve the road rutting detection by disregarding any false positive prediction made by the model which are not on the road surfaces.

Further, this is the first time that detection of road damage focusing solely on road rutting has been done using deep learning. This has several advantages. Firstly, since there is no previous study on the detection of road rutting using deep learning on images, the results obtained in our study can be used as a benchmark for similar works in future. Secondly, the proposed new road rutting dataset can be combined with dataset containing other road damage types to train road damage detection models. This will help assign road rutting a unique category rather than merging it into a category with other road damage types. Inclusion of more road rutting images will help improve its detection accuracy which otherwise was not possible as remarkably less instances of road rutting was included in the past research. This will help develop a better model for road damage detection for practical applications  as road rutting is a severe road damage type besides cracks, and potholes, and its identification as a separate category is important to enable road managers to take effective remedial actions. This will be helpful in giving boost to collection of more road rutting dataset as majority of the present researches focus more on collection of dataset containing images of cracks and potholes.

\section{conclusion}
\label{sec:conclusions}
In our research, we developed a novel road rutting dataset containing 949 images from heterogeneous sources. We used the state of the art object detection models, namely YOLOv4, YOLOv5, YOLOv6, and YOLOX, and semantic segmentation models, namely PSPNet (Resnet-50) and DeepLabv3+ (Resnet-101) to study the feasibility of detection of road rutting using deep learning. We performed quantitative and qualitative analysis of the models and their prediction to evaluate their performance, make comparisons, and to identify challenges faced in performing this task using the proposed method. Object detection model YOLOX-s achieves mAP@IoU=0.5 of 61.6\% and semantic segmentation model PSPNet (Resnet-50) achieves IoU of 54.69 and accuracy of 72.67. Besides quantitative results, visual or qualitative analysis of model predictions helped find out that deep learning models used in the study provided stable predictions in validating images.

In continuation of the research work, we want to further increase the road rutting dataset and combine it with the already existing road damage dataset containing cracks, potholes, etc. to develop a comprehensive model for road damage detection.

\bibliographystyle{unsrt}
\bibliography{mybib}

\end{document}